# Anomalous Situation Detection in Complex Scenes

Michalis Voutouris, Giovanni Sachi, Hina Afridi
University of Trento, Italy

Abstract— *In this paper we investigate a robust method to identify anomalies in complex scenes. This task is performed by evaluating the collective behavior by extracting the local binary patterns (LBP) and Laplacian of Gaussian (LoG) features. We fuse both features together which are exploited to train an MLP neural network during the training stage, and the anomaly is identified on the test samples. Considering the challenge of tracking individuals in dense crowded scenes due to multiple occlusions and clutter, in this paper we extract LBP and LoG features and use them as an approximate representation of the anomalous situation. These features well match the appearance of anomaly and their consistency, and accuracy is higher both in regular and irregular areas compared to other descriptors. In this paper, these features are exploited as input prior to train the neural network. The MLP neural network is subsequently explored to consider these features that can detect the anomalous situation. The experimental tests are conducted on a set of benchmark video sequences commonly used for anomaly situation detection.*

Keywords; Anomalous situation; Laplacian of Gaussian; Neural Network; Binary Patterns;

I. INTRODUCTION

The gatherings of public at various famous events present problems of paramount significance to public safety and security [1][2][3]. These environments are dangerous in the existence of anomalous situations in terms of riots and chaotic acts of crowd as a whole in populous areas [4][5]. It is significant to identify the happening of anomalies early because it can alleviate the potential dangerous results and can also warn a human operator for monitoring the ongoing environment more attentively. However, the monitoring of video generally entails human operators to keep an eye on the display screens, which often leads to fatigue, inattention, and failure to detect the happening of abnormal events [6][7]. On the other hand, important problems arise with the substantial amount of surveillance video data, which are difficult and time-consuming for manual analysis [8][9][10]. Considering these challenges, an automated anomaly detection system gains increasing interest from both academia and industry.

Early research related to anomaly detection mainly focuses on a particular set of tasks. For instance, Rota et al. [25] exploited influence of particles to identify different groups, and Ullah et al. [11][12] explored a different method to identify abnormal behavior. However, general video anomaly detection presents challenges due to the vague explanation of anomaly in practical situations. One may consider that the existence of a vehicle on a pedestrian pathway is normal, but others may think of it an anomaly. In fact, an anomaly is an observation that does not have consistency with other observations over time. Considering the inconsistency between normal and abnormal behaviors, one can model normal patterns in an unsupervised or semisupervised setup, and the pattern that do not correlate with the model is considered as anomaly. Significant research has been conducted on feature modeling for normal patterns. In sub-urban areas where classic target tracking can be well designed, high-level features, such as corner features, can be exploited for anomaly detection. However, in cities where not all targets can be accurately tracked, low-level features including histogram of oriented optical flow and gradients, social force models [3][13], and motion scale-invariant feature transform are robust for extraction, are often explored to detect anomalous events in videos.

Apart from that, the strength of these features for addressing other problems, they are easily influenced by the background. Thus, the aforementioned features cannot focus on the object of interest, for instance, anomaly. In this paper, we model a unique method that fuses widely tested features for different computer vision problems. In our paper, we fuse Laplacian of Gaussian and local binary patterns. These features are extracted from individual images. Then we model a feature vector concatenated both the features. The feature vector is fed to MLP neural network that identifies anomaly in diverse videos. The rest of the paper is divided into related work, proposed method, experiments, and conclusion.

## II. RELATED WORK

In the methods proposed in [14][15], crowded situations is considered as a evolving flow field, which is the famous method nowadays. Saqib et al. [16] and the researchers in [17][18][19] paid attention to the extraction of crowd attributes from the vector fields to demonstrate crowd density, moving directions, and boundaries. In recent years, the focus has shifted towards application-oriented methods to enhance crowd pattern interpretation [20][21][22]. Ullah et al. [23][24][25] first investigated a crowd scene model using the optical flow field - an extension of the flow-filed model - for segmenting extremely populous crowd scenes in videos. This method has also been exploited in group tracking that consists of a lot of or intersected crowd entities [26]. Khan et al. [28] off-line crowd moving direction learning method [27] has also been significantly effective flow-based tracking method. Crowd Kanade-Lucas-Tomasi corners, multi-label optimization and Lagrangian modeled anomaly crowd visual features from flow field information. Those models described their potentials in tracking the dynamic crowd under extremely crowded and partial occluded situations but are bound to pre-defined crowd patterns.

Some models present the crowd scene videos as Spatio-temporal Volume (STV), which fuses global video dynamics into a three-dimensional feature space. For instance, [29][30] investigate a motion labelling model based on the co-occurrence of features. The model is created as a potential function in the Markov Random Field (MRF) process for anomaly detection. The techniques in [31][32] proposed STV-based motion patterns in volumetric environment to represent the spatial-temporal statistical characteristics of pedestrians in crowded environments. The techniques [33][34][35][36] modeled a STV-based anomaly location detection method exploiting localised cuboids in an unsupervised learning way. These techniques are valid approaches when using spatio-temporal features without parameter configurations. Some methods [37][38] use deep learning based techniques for event detection.

## III. PROPOSED METHOD

The Laplacian is a way of the second spatial derivative of a video frame. The Laplacian of a frame presents regions of rapid intensity change and is therefore often exploited for feature extraction. The Laplacian is generally explored in connection with a Gaussian smoothing filter to smooth the video frame and to decrease its sensitivity to different noises. The Laplacian of the Gaussian (LoG) is a very famous and effective method for extracting features from the frames. Provided an input frame f(x,y), this frame is executed by a Gaussian kernel as modeled in Eq. (1).

$$G(x,y,t) = \frac{1}{2\pi t} e^{-\frac{x^2+y^2}{2t}} \qquad (1)$$

The outcome of LOG is achieved according to the provided Eq. (2).

$$L(x,y,t) = G(x,y,t) * f(x,y) \qquad (2)$$

The LOG sometime results in strong positive responses for dark areas of a specific radius and strong negative responses for bright areas of different sizes. An important problem when using LOG at a single scale is that the outcome is strongly dependent on the relationship between the size of the area structures in the video frame and the size of the Gaussian kernel exploited for pre-smoothing. In order to automatically encode anomalous locations of various sizes in the frame domain, a multi-scale method is therefore important.

An easy method to achieve a multi-scale location detector with automatic scale selection is to consider the scale-normalized Laplacian extractor and to identify scale-space maxima/minima, that are pixel points in accordance with both space and scale. Exploiting the provided input frame, we find scale-space volume. We identify a point as a bright if the value at this point is greater than the value in all its neighbors. This technique gives a precise operational definition, which directly leads to an efficient and effective model for feature extraction. The significant characteristics from scale-space maxima of the normalized Laplacian technique are the covariant responses with translations, rotations and rescaling in the frame under investigation. It is important that if a scale-space maximum is used at a point under a rescaling of the frame by a scale factor, there will be a scale-space maximum at in the rescaled frame. This feature means that apart from the specific topic of Laplacian point detection, local maxima/minima of the scale-normalized Laplacian are also exploited for scale selection in other contexts, such as in corner detection, scale-adaptive feature tracking, scale-invariant feature transform as well as other image descriptors for image matching and object recognition.

To improve the effectiveness of the extracted feature, we combine local binary patterns (LBP) with LOG. The LBP is a type of visual descriptor exploited for classification in vision problems. It is a strong feature for texture classification; we have investigated that when LBP is integrated with the LOG, it enhances the anomaly detection performance significantly. We design the LBP feature vector in a few easy stages. We categorize the examined window of a video frame into cells and for each pixel in a cell, we compare the pixel to each of its 8 neighbors. We record 0 where the center pixel's value is greater than the neighbor's value, otherwise, we record 1. This gives an 8-digit binary pattern. We then calculate the histogram, over the cell, of the frequency of each number happening (i.e., each combination of which pixels are smaller and which are greater than the center). Therefore, this histogram can be seen as a 256-dimensional feature vector. We normalize the histogram and integrate histograms of all cells. This gives a feature vector for the entire window.

We decrease the length of the feature vector and code a simple rotation invariant descriptor because some binary patterns happen more commonly in anomaly frames than others. A local binary pattern is called uniform if the binary pattern consist of at most two 0-1 or 1-0 transitions. For example, 00010000 (2 transitions) is a uniform pattern, and 01010100 (6 transitions) is not a uniform pattern. In the computation of the histogram, the histogram has a isolate bin for every uniform pattern, and we assign all non-uniform patterns to a single bin. Using uniform patterns, the length of the feature vector for a single cell decreases from 256 to 59. The 58 uniform binary patterns correspond to the integers 0, 1, 2, 3, 4, 6, 7, 8, 12, 14, 15, 16, 24, 28, 30, 31, 32, 48, 56, 60, 62, 63, 64, 96, 112, 120, 124, 126, 127, 128, 129, 131, 135, 143, 159, 191, 192, 193, 195, 199, 207, 223, 224, 225, 227, 231, 239, 240, 241, 243, 247, 248, 249, 251, 252, 253, 254 and 255.

We model the feature vector using neural network to identify anomalies. We fuse both features, and train an MLP neural network. The motivation for using MLP is in its substantial ability, through backpropagation, to hinder noise, and the dexterity to generalize. The extracted features are fed as an input to the MLP. The result is obtained by spreading the features as an input vector through the hidden layers. In MLP networks, there are L + 1 layers of neurons, and L layers of weights. During the training stage, the weights W and biases b are modified so that the actual output becomes closer to the desired output. For this purpose, a cost function is presented as in Eq. (3).

$$E(W,b) = \frac{1}{2}\sum_{i=1}^{n_l}(d_i - y_i^L)^2 \quad (3)$$

This function computes the squared error between the desired and actual output vectors and the backpropagation is gradient descent on the cost function in Eq. (3). Therefore, during the training stage, weights and biases modifications are equivalent to Eq. (4) and Eq. (5), respectively.

$$\Delta W_{ij}^{l} = -\eta \frac{\partial E}{\partial W_{ij}^{l}} \quad (4)$$

$$\Delta b_{ij}^{l} = -\eta \frac{\partial E}{\partial b_{ij}^{l}} \quad (5)$$

The backpropagation method starts with the forward pass where the input vector is transformed into output. The difference between the desired output and the actual output is computed to estimate the error. During the backward pass, the estimated error at the output units is propagated backwards through the network. The weights and biases are updated, for l = 1 to L using the results of the forward and backward passes. The learned weights and biases from the training stage are exploited to identify anomaly.

IV. EXPERIMENTS

We analyze the performance of our model for anomalies detection on UMN dataset available publicly. The UMN dataset comprises of normal and abnormal crowd videos from the University of Minnesota. It includes three different indoor and outdoor scenes representing 11 different scenarios of escape events. There are total 7739 frames of 320x240 pixels. Each video starts with the normal behaviors of people walking and standing. There are also students moving across two buildings lasting for 12 and 5 minutes, respectively. Each sequence is segmented into two different subsequences with people mainly moving in a horizontal direction in the scene. This dataset presents anomaly as the deviations from what has been observed beforehand. This anomaly represents any pedestrian moving in the opposite direction of the general flow of the pedestrians. The detection of anomalies in terms of panic situation is shown for the four sample videos in the Figure.

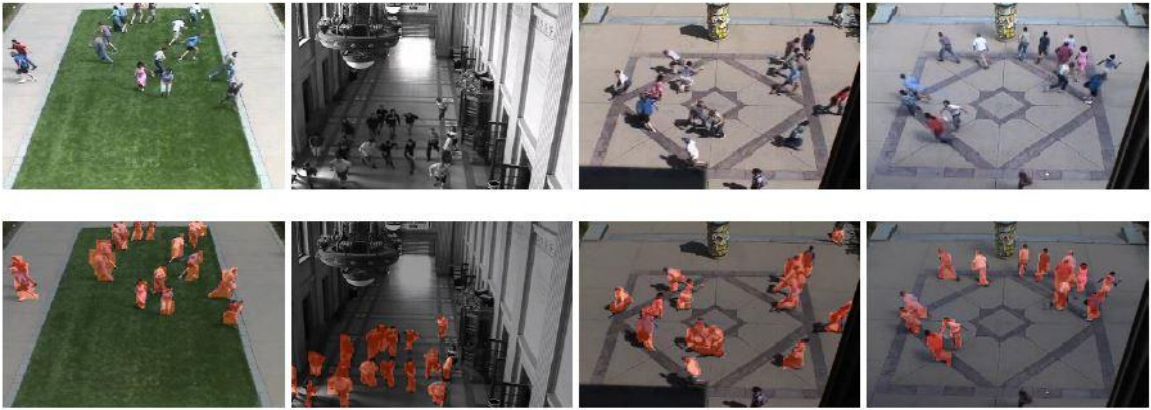

Figure 1. UMN Dataset: Four sample frames from four videos are shown in the first row. The second row shows the results.

We also showed quantitative results in term of F-scores for the four sequences of UMN dataset in Table 1 below. We obtained very good performance.

TABLE 1. F-scores for the UMN dataset are presented.

| Dataset | Sequences | F-scores |
|---------|-----------|----------|
| UMN     | *S1*      | *0.56*   |
|         | S2        | 0.78     |
|         | S3        | 0.66     |
|         | S4        | 0.71     |
| Average |           | 0.67     |